# Advanced Integration of Discrete Line Segments in Digitized P&ID for Continuous Instrument Connectivity

Soumya Swarup Prusty
swarup.samrat@gmail.com

Astha Agarwal
asthaa29@gmail.com

Srinivasan Iyenger

## ABSTRACT:

Piping and Instrumentation Diagrams (P&IDs) constitute the foundational blueprint of a plant, depicting the interconnections among process equipment, instrumentation for process control, and the flow of fluids and control signals. In their existing setup, the manual mapping of information from P&ID sheets holds a significant challenge. This is a time-consuming process, taking around 3-6 months, and is susceptible to errors. It also depends on the expertise of the domain experts and often requires multiple rounds of review. The digitization of P&IDs entails merging detected line segments, which is essential for linking various detected instruments, thereby creating a comprehensive digitized P&ID. This paper focuses on explaining how line segments which are detected using a computer vision model are merged and eventually building the connection between equipment and merged lines. Hence presenting a digitized form of information stating the interconnection between process equipment, instrumentation, flow of fluids and control signals. Eventually, which can be stored in a knowledge graph and that information along with the help of advanced algorithms can be leveraged for tasks like finding optimal routes, detecting system cycles, computing transitive closures, and more. [Reference 2]

## 1. INTRODUCTION:

Piping and Instrumentation Diagrams (P&IDs) play a crucial role in the process plant industry, serving as essential design documents during the equipment, the instrumentation systems responsible for process control, and the pathways for fluid and control signal flow. Additionally, P&IDs act as foundational reference material throughout various project phases, including detailed engineering, procurement, construction, and plant commissioning.

Digitization of P&ID diagram enables easier access, faster updates, improved collaboration, and streamlined workflows across an industrial facility, ultimately leading to increased operational efficiency, reduced maintenance downtime, and better decision-making by providing readily available, accurate information about the plant's processes and equipment.

To convert an image format P&ID into a digital P&ID, a process for recognizing high-level objects in a related application field from the image diagram, extracting the necessary information to reconstruct connection relations between the objects, and creating a structured diagram is required to generate a computer interpretable P&ID.

A digital P&ID is represented such that all objects in the diagram are structured and can be digitally processed. The major components of a digital P&ID are symbols that can be divided into equipment, piping, and instrumentation. The symbols in a digital P&ID have connecting lines that are distinguished as piping or signal lines. [Reference 1]

This study focuses on the aspect of digitization process that involves converting separate piping segments into interconnected piping segments. In a digitized P&ID, continuous piping is represented as precise, data-rich objects that link different symbols—such as valves, pumps, tanks, and instruments—within the system. And it provides very comprehensive information to the end user.

Integration of continuous piping segments into digital systems allows for automatic validation, error checking, and updating of diagrams. By focusing on digitizing the continuous piping segments that connect various system components, companies can pave the way for more intelligent and automated plant operations, improving both the efficiency and reliability of industrial processes.

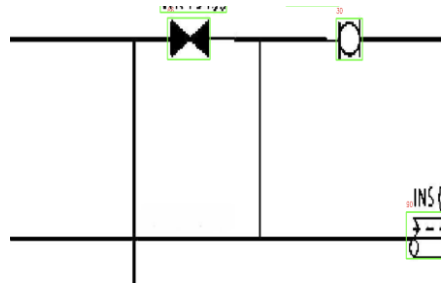

Figure 2: Symbol detection output

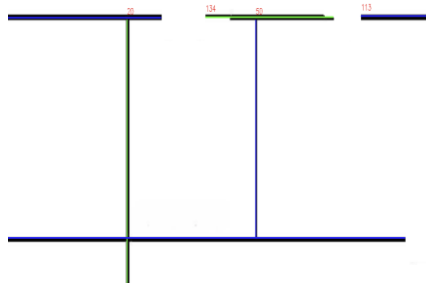

Figure 3: Line detections output

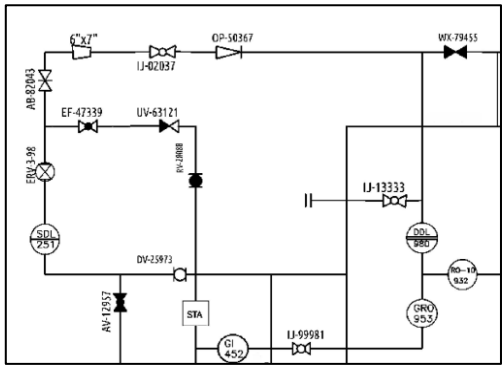

Figure 1: Standard snippet of P&ID Diagram

## 2. PROBLEM STATEMENT:

Once the instruments/objects are identified (see Figure 2) and line segments are detected (see Figure 3), the need is to combine the information, i.e. to merge line segments to build line continuity (see Figure 4) such that instrument interconnections to other instruments, such as valves, tank, pump can be derived out (see Figure 5).

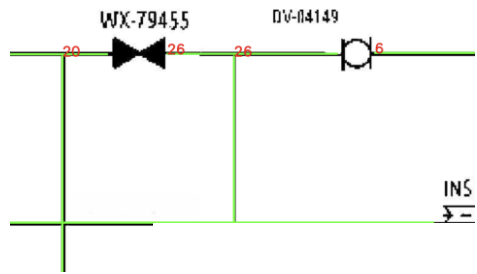

Figure 4: Related lines across symbols

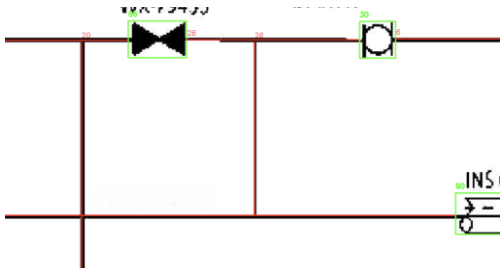

Figure 5: Identify interconnections between the symbols

## 3. PRE-REQUISITES:

1. We have maintained a prerequisite that P&ID diagram should be with 200 dpi or more. If not by default, then we convert the P&ID diagram.
2. P&ID diagram is properly aligned; no extra effort is taken to align the diagram.
3. While detecting line, capturing the direction of liquid flow is not considered.

## 4. METHODOLOGY:

Before merge segmentation, digitization of P&ID involves multiple steps. [Reference 8] [Reference 3] [Reference 4]
1. It starts with object detection, where it identifies all the equipment as symbols.
2. Next in the object classification step the detected symbols are classified. The purpose is to find out the detected symbol types that are identified in the earlier step.
3. Meanwhile all the texts in the P&ID diagram are read and they are mapped to classified symbols based on vicinity. Texts provide more context in further classifying the symbols.
4. Next, we identify all line segments (vertical and horizontal) in the P&ID diagram. These lines are nothing but piping instruments. And these discrete line segments and the classified symbols are mapped together based on vicinity and connection.

Post all the above-mentioned steps we have discrete/non-connected lines. So, to carve out meaningful information from this state, we need to connect the lines wherever necessary to build the connection between instruments. This process is called merged line segments.

Before getting into merge segments, it's noteworthy that in line detection step, at times because of width of line segment(s), we observed multiple parallel lines being detected instead of just one line segment. We handled that duplication detection scenario, which helped us improve our accuracy.

Step 1:
In this step the algorithm looks for edge to edge vicinity, in this step line segments whose edges are close to each other (see figure 6a), either vertically or horizontally, are clubbed into a single key value pair (see figure 6b). Likewise, the algorithm traverses through all line segments and performs the above operation and eventually all line segments are presented by a key value pair. Where the key is the new merged line number and list representing individual segments [Reference7] [Reference 5].

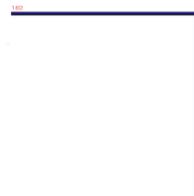

Figure 6a: Representation before merging segments

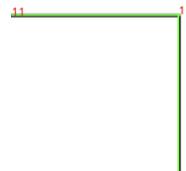

Figure 6b: Representation after merging segments

**Step 2:**
In this step the algorithm looks for a scenario where one edge meets another line. In this the connection is most likely perpendicular (see figure 7a) and if the edge of one segment rests on another segment, then they clubbed into single key value pair (see figure 7b). While doing so the algorithm avoids line segments crossing each other (see figure 7c), here line 20 and 26 are intersecting each other and hence they are part of single key pair value

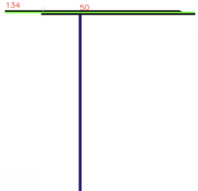

Figure 7a

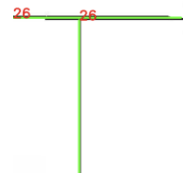

Figure 7b

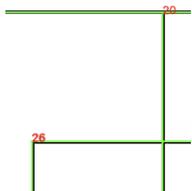

Figure 7c

**Step 3:**
After step 2, there will be a resultant dictionary of key value pairs. Now the algorithm will traverse through all value lists to find if any line segment is common across multiple keys and if it finds then it merges those values into a single key and previous keys for which it encounters this scenario are eliminated.

**Step 4:**
As a clean-up process, we delete any line segment which is not connected to any of the instrument/object.

Once this step is over, we have the final dictionary (see figure 10) where the key represents a new unique line number (see figure 8), and the list includes all the line segments that constitute this resultant line (see figure 9).

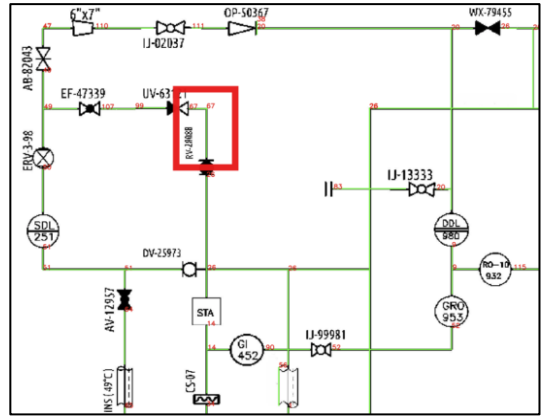

Figure 8: Standard snippet of P&ID Diagram with merged line segments

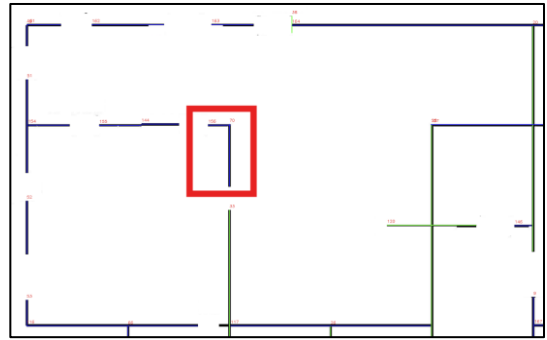

Figure 9: Snippet of P&ID Diagram with only detected line segments

```
{
    67:[156,70],
    49:[51,154],
    51:[53,116,66]
}
```

Figure 10: Snippet of final dictionary

## 5. EXPERIMENTS:

For our experiments and evaluation, we used sample P&ID diagrams from DigitizePID_Dataset - Google Drive [Reference 6] repository.

We selected 30 P&ID diagrams with a mix of easy, medium and complex P&ID diagrams, where we maintained a resolution of 200dpi across all diagrams.

## 6. RESULTS:

Post our algorithm runs on sample data set, below metrices (see table 1) show the efficiency.

| F1 Score | Recall | Accuracy |
|---|---|---|
| 96.2% | 98.5% | 93.65% |

Table 1: Algorithm output metrices

We have achieved an accuracy of 93.65%, compared to the previously recorded 90.6% [Reference 2].

The following metrics were obtained when we applied the aforementioned algorithm to highly complex and confidential P&ID diagrams.

| F1 Score: 85% | Recall: 85% |
|---|---|

Table 2: Metrices on confidential complex data

## FUTURE SCOPE:

Traditional P&ID diagrams have 'line break'. Which means even if there is a continuous line segment but if it has line break attached to it, then the definition of line segment changes. As each line represents a connection between equipment and each connection has a connection identification number. But if there is a presence of line break then the connection would have more than one identification number.

P&ID diagrams are predominantly used in industrial settings particularly in process industries like chemical manufacturing, oil refineries, paper mills, and power plants.